\documentclass[10pt,twocolumn,letterpaper]{article}

\usepackage[pagenumbers]{cvpr}            
\usepackage{tabularx}
\usepackage{multirow}
\definecolor{cvprblue}{rgb}{0.21,0.49,0.74}
\usepackage[pagebackref,breaklinks,colorlinks,allcolors=cvprblue]{hyperref}

\def\paperID{18} 
\def\confName{CVPR}
\def\confYear{2026}

\title{From Memorization to Creativity: \\ LLM as a Designer of Novel Neural Architectures}

\author{Waleed Khalid  \hspace{2cm} Dmitry Ignatov$^*$  \hspace{2cm} Radu Timofte\\
{\small Computer Vision Lab, CAIDAS \& IFI, University of W\"urzburg, Germany $\cdot$ $^*$dmytro.ignatov@uni-wuerzburg.de}}

\begin{document}

\maketitle

\begin{abstract}
Large language models (LLMs) excel in program synthesis, yet their capacity for neural architecture design---balancing syntactic reliability, performance, and structural novelty---remains underexplored. We present a closed-loop architecture synthesis pipeline within the NNGPT framework, in which a code-oriented LLM evolves over 22 supervised fine-tuning cycles. At each cycle, the LLM synthesizes PyTorch convolutional networks, validated via low-fidelity performance signals and filtered via a MinHash--Jaccard criterion to prevent structural redundancy before being incorporated into the LEMUR dataset. High-performing candidates with novel architectures are converted into prompt--code pairs for parameter-efficient LoRA fine-tuning. This feedback loop drives a measurable distributional shift, progressively internalizing empirical architectural priors such that valid and high-performing outputs evolve from scarce to dominant across cycles. On CIFAR-10, the valid generation rate stabilizes at 50.6\% (peaking at 74.5\%), mean first-epoch accuracy rises from 28.1\% to 51.0\%, and candidates exceeding 40\% accuracy grow from 2.0\% to 96.8\%. Cross-dataset transfer to CIFAR-100 and SVHN confirms that improved validity, shifted accuracy distributions, and sustained novelty generalize across benchmarks of varying difficulty and visual domain. Across 22 cycles, 455 unique architectures absent from the original corpus are admitted under the novelty filter. By grounding synthesis in execution feedback and novelty filtering, we demonstrate that iterative self-supervised fine-tuning reshapes an LLM into a task-specialized architectural prior---improving generation reliability, proxy performance, and structural diversity---offering a reproducible, annotation-free alternative to hand-crafted search spaces. 
All code artifacts are publicly released under {\small \url{https://github.com/ABrain-One}}, including the code retrieval engine (\href{https://github.com/ABrain-One/nn-rag}{NN-RAG}), prompt generation (\href{https://github.com/ABrain-One/nn-dup}{NN-Dup}), and LLM fine-tuning (\href{https://github.com/ABrain-One/nn-gpt}{NNGPT}) pipelines, as well as the generated models (\href{https://github.com/ABrain-One/NN-Dataset}{NN Dataset}); the fine-tuned LLM is available at {\small\url{https://huggingface.co/ABrain/NNGPT-UniqueArch-Rag}}.
\end{abstract}
    
\section{Introduction}
Designing effective neural network architectures remains a central bottleneck in modern deep learning. Neural Architecture Search (NAS) emerged to automate this process through reinforcement learning, evolutionary algorithms, and differentiable optimization \citep{zoph2017neural, white2023neural, kang2023neural}. While successful, traditional NAS often incurs prohibitive computational costs. Consequently, the CIFAR-10 classification task \citep{krizhevsky2009learning} has become a canonical benchmark for evaluating these automated design strategies.

In parallel, large language models (LLMs) have revolutionized program synthesis, enabling the generation of complex source code from natural-language instructions. Recent frameworks like \textit{LLMatic} and \textit{LEMONADE} have begun leveraging LLMs to emit full network definitions, demonstrating their potential as architecture generators \citep{nasir2024llmatic, rahman2025lemonade}. However, existing studies primarily focus on final model accuracy and search efficiency, offering limited insight into the generator's reliability. Specifically, it remains unclear how LLM-driven synthesis evolves under iterative refinement, particularly regarding syntactic \emph{validity}, structural \emph{novelty}, and the ability to maintain diversity as the model specializes.

In this work, we therefore consider an LLM purely as an \emph{architecture synthesizer} and address the following central question: if we repeatedly fine-tune an LLM on its own successful generations, does its ability to produce valid, high-quality, and structurally novel network architectures measurably improve over time? Rather than optimizing for final test accuracy after long training runs, we deliberately adopt a low-cost performance proxy: the classification accuracy achieved after a \emph{single} training epoch on CIFAR-10 \citep{krizhevsky2009learning}. This early-epoch accuracy is inexpensive to obtain and directly reflects how well the generated architectures support fast initial learning. At the same time, we treat structural uniqueness as a first-class objective, because practical NAS workflows benefit not only from strong individual models but also from diverse candidates that explore different regions of the design space \citep{white2023neural,kang2023neural}.

Concretely, we execute an LLM-driven synthesis loop over 22 cycles where candidate architectures are filtered for compilation validity, trained for a single epoch, and subjected to MinHash--Jaccard novelty analysis. Our results demonstrate that this iterative generate--evaluate--select--fine-tune process, guided by low-fidelity signals, produces a pronounced upward shift in the first-epoch accuracy distribution across cycles, while maintaining significant structural diversity as measured by code-level novelty filtering. The valid generation rate, though not monotonic, stabilizes at levels substantially above early-cycle baselines for much of the run, effectively reshaping the LLM into a task-specialized architectural prior.

It is important to delineate the scope of this study: we do \emph{not} propose a complete NAS method competing on final test accuracy against established search algorithms, nor do we claim that the generated architectures surpass hand-designed baselines after full training. Instead, we study \emph{the generator itself}---how its output distribution over architectures changes under iterative self-refinement---using first-epoch accuracy as a low-cost, interpretable proxy. The contribution is therefore a characterization of LLM-as-generator dynamics, providing a foundation upon which future work can build by coupling the refined generator with downstream optimization.

In summary, this work advances the intersection of automated program synthesis and neural architecture design through four primary contributions:
\begin{enumerate}
    \item We establish an LLM-driven synthesis framework that treats the generator as a trainable architectural prior, optimizing for a triad of objectives: syntactic validity, early-epoch performance, and structural novelty.
    \item We introduce a code-level novelty filter utilizing MinHash--Jaccard similarity to programmatically ensure meaningful design-space expansion.
    \item We provide a 22-cycle longitudinal analysis demonstrating that iterative fine-tuning induces a substantial distributional shift toward higher-performing architectures, with net improvements in generation reliability, without sacrificing architectural diversity.
    \item We validate the generality of the observed trends across three benchmarks---CIFAR-10, CIFAR-100, and SVHN---showing that the iterative refinement loop induces consistent distributional shifts regardless of task difficulty, number of classes, or visual domain.
\end{enumerate}

\section{Related Work}
\label{sec:related-work}

The development of NAS has significantly automated network design through reinforcement learning, evolutionary algorithms, and differentiable optimization, though often at a prohibitive computational cost \citep{elsken2019nas,white2023neural,kang2023neural}. To ameliorate the expense of repeated candidate training, the field has increasingly relied on low-fidelity proxies, including early-stopped training, learning-curve extrapolation, and training-free zero-cost signals \citep{domhan2015lcextrapation,ru2020speedy,zela2020darts}. While our work utilizes single-epoch accuracy as an efficient performance proxy, we diverge from traditional NAS by employing these signals to shape the behavioral priors of a generative LLM rather than optimizing within a static, handcrafted search space.

In parallel, the advent of code-capable LLMs has introduced a paradigm shift toward synthesizing complete model implementations from natural language. Frameworks such as \textit{LLMatic} have demonstrated the efficacy of coupling LLM-driven mutation with quality-diversity search \citep{nasir2024llmatic}, while others have integrated iterative refinement to satisfy stringent deployment constraints \citep{rahman2025lemonade}. More recent self-improving systems, such as \textit{SEKI} and \textit{RZ-NAS}, leverage performance-guided evolution or reflective reasoning to improve design outcomes \citep{cai2025seki,ji2025rznas}. Despite these advances, existing research typically evaluates success through final search efficiency or peak accuracy. Our approach provides a distinct longitudinal perspective by explicitly characterizing the evolution of the generator itself---tracking metrics of validity, performance distribution shifts, and code-level novelty across twenty-two successive cycles of supervised fine-tuning.

Crucially, the utility of LLM-generated architectures is predicated on both functional reliability and structural diversity. While standard code generation benchmarks prioritize functional correctness and unit-test pass rates \citep{chen2021codexeval,jiang2024codellmsurvey}, the structured nature of PyTorch programs necessitates a more nuanced separation between executable validity---comprising parsing, instantiation, and forward passes---and downstream learning quality. To prevent the collapse of the generator into redundant motifs or trivial rewrites, we incorporate a MinHash--Jaccard near-duplicate filter, ensuring that the fine-tuning corpus is augmented only with implementations that are both performant and structurally novel. Table~\ref{tab:method_comparison} summarizes the key methodological and empirical distinctions between our framework and the most closely related LLM-based architecture generation methods.

\begin{table*}[htbp]
  \centering
  \small
  \renewcommand{\arraystretch}{1.15}
  \begin{tabularx}{\textwidth}{@{}l>{\centering\arraybackslash}X>{\centering\arraybackslash}X>{\centering\arraybackslash}X>{\centering\arraybackslash}X>{\centering\arraybackslash}X>{\centering\arraybackslash}X@{}}
    \toprule
    \textbf{Method} & \textbf{\mbox{Valid Rate}} & \textbf{\mbox{Proxy Metric}} & \textbf{\mbox{Novel Filter}}
      & \textbf{\mbox{Iter.\ FT}} & \textbf{\mbox{Corpus Growth}} & \textbf{\mbox{Best Accuracy}} \\
    \midrule
    LLMatic~\citep{nasir2024llmatic}
      & N/R & Final & QD arch. & No & No & ${\sim}94\%^{a}$ \\
    LEMONADE~\citep{rahman2025lemonade}
      & ${\approx}75$--$85\%^{\dagger}$ & Constr. & None & No & No & N/A$^{b}$ \\
    SEKI~\citep{cai2025seki}
      & N/R & Final & Repo. & Yes & N/R & $97.7\%^{a}$ \\
    RZ-NAS~\citep{ji2025rznas}
      & ${\approx}98\%^{\ddagger}$ & Zero-cost & None & No & No & N/R$^{c}$ \\
    \midrule
    \textbf{Ours}
      & \textbf{50.6\%}$^{e}$
      & \textbf{1-ep.}
      & \textbf{MinHash}
      & \textbf{Yes}
      & \textbf{Yes}$^{f}$
      & \textbf{64.0\%}$^{d}$ \\
    \bottomrule
  \end{tabularx}
  \caption{Comparison of LLM-based architecture generation frameworks
  (CIFAR-10). N/R\,=\,not reported; FT\,=\,fine-tuning.
  $^{a}$\,Fully-trained test accuracy (DARTS/NB201 search space).
  $^{b}$\,Optimizes deployment constraints, not peak accuracy.
  $^{c}$\,Uses zero-cost ranking; final accuracy not reported.
  $^{d}$\,\textbf{First-epoch validation accuracy only}; not directly
  comparable to fully-trained results above.
  $^{e}$\,Explicitly tracked across 22 cycles; no other method reports longitudinal validity.
  $^{f}$\,Training corpus grows from 1{,}698 to 2{,}153 prompt--code pairs across cycles.
  Our framework is the only one to jointly track valid generation rate,
  apply a code-level novelty filter, grow the training corpus iteratively,
  and fine-tune on low-fidelity proxy feedback---all without manual annotations.}
  \label{tab:method_comparison}
\end{table*}

\section{Method}
\label{sec:method}

\begin{figure}[htbp]
  \centering
  \includegraphics[width=\linewidth]{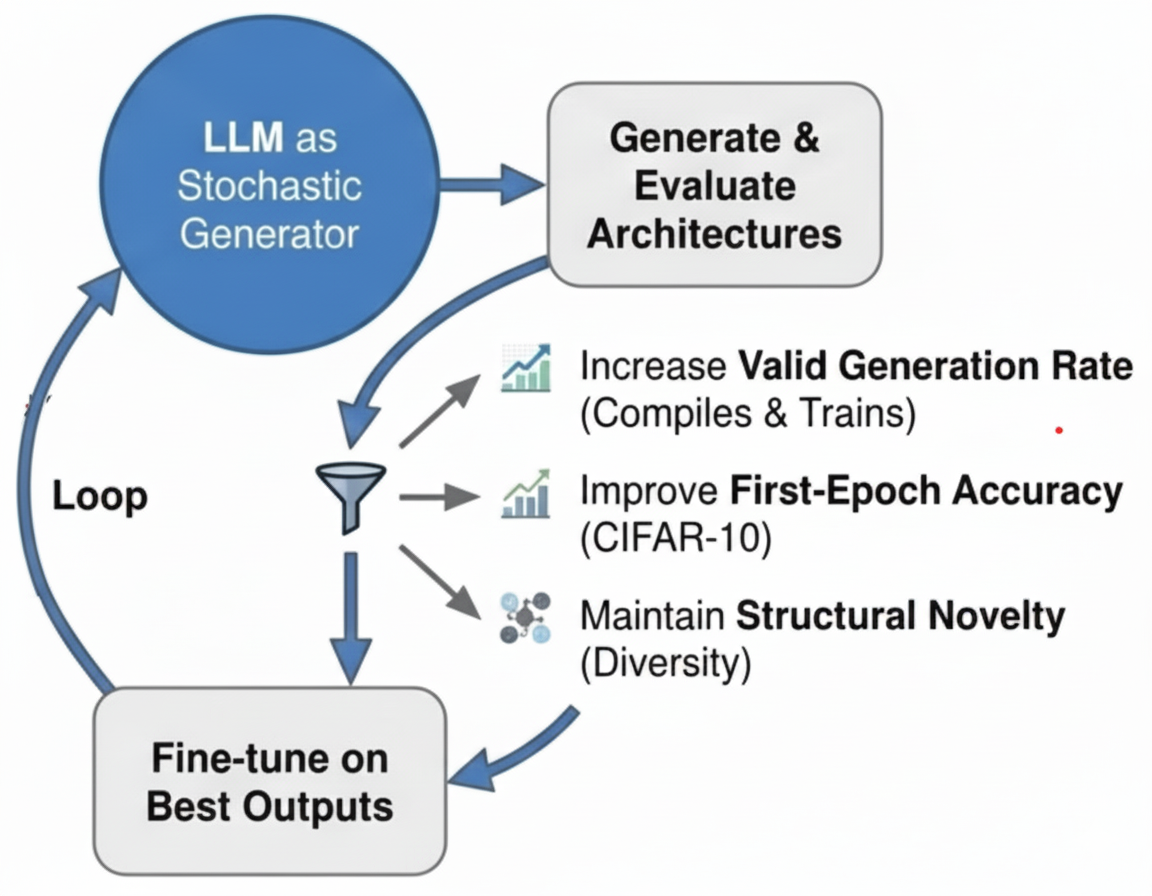}
  \caption{Overview of the iterative architecture-synthesis loop: the LLM generates architectures, candidates are evaluated and filtered (validity, first-epoch accuracy, novelty), and the LLM is fine-tuned on selected outputs.}
  \label{fig:generator}
\vspace{-5mm}
\end{figure}

We treat a code-oriented large language model (LLM) as a stochastic generator of neural network architectures and study how its behavior changes under an iterative refinement loop. We run $22$ synthesis cycles indexed by $c \in \{1,\dots,22\}$. In each cycle, the LLM generates candidate PyTorch models; we execute validity checks, run a fixed first-epoch training protocol to obtain a low-cost performance proxy, filter for novelty, and then fine-tune the LLM on the accepted outputs before proceeding to the next cycle. The primary evaluation is conducted on CIFAR-10~\citep{krizhevsky2009learning}; cross-dataset generalization is assessed on CIFAR-100~\citep{krizhevsky2009learning} and SVHN~\citep{netzer2011svhn} using the same loop with dataset-appropriate accuracy thresholds (Section~\ref{sec:cross_dataset_setup}). Figure~\ref{fig:generator} summarizes the generate--evaluate--select--fine-tune loop.

A code-focused LLM is used as a conditional sampler over PyTorch architectures. Prompts specify CIFAR-10 classification, input/output shapes ($(N,3,32,32)\rightarrow 10$ logits), and constraints on permissible operations (standard conv/pool/norm/activations; no pretrained weights). A parameter budget of at most $500{,}000$ parameters is imposed. Each generated model must implement a fixed API contract (class \texttt{Net(nn.Module)} with \texttt{forward}, \texttt{train\_setup}, \texttt{learn}); the full prompt specification is provided in Suppl.~\S I. The prompt template, decoding configuration, and maximum generation length are held fixed across all cycles to avoid prompt drift; consequently, changes in outputs are attributable to training data and fine-tuning.

We initialize supervision using the LEMUR Neural Network Dataset~\citep{ABrain.NNGPT,ABrain.NN-Dataset}, a broad collection of performance-annotated neural network implementations spanning high-capacity and edge-optimized designs~\citep{ABrain.LEMUR2,ABrain.NN-Lite,ABrain.MobileAgeNet,ABrain.MobileDenoising}, drawing on prior work in LLM-driven architectural synthesis within the NNGPT ecosystem~\citep{ABrain.NN-RAG,ABrain.HPGPT,ABrain.NN-Captioning_2025,ABrain.Prompt,ABrain.NNGPT-Fractal,ABrain.Transform,ABrain.CV_Channel,ABrain.Feedback_Memory}. After MinHash/LSH deduplication and chat-format conversion (Suppl.~\S A--C), the corpus yields 849 unique architectures paired with two task descriptions, producing 1{,}698 prompt--code training examples. 

For every successfully parsed and trained candidate, we compute two Jaccard similarities over token-level shingles of the model code. The first, $J_{\text{train}}$, measures similarity to the deduplicated supervised training set; the second, $J_{\text{gen}}^{(c)}$, measures similarity to the set of all code samples generated earlier in the same cycle $c$. Formally, let $S_{\text{train}}$ denote the collection of shingle sets corresponding to all models in the current supervised corpus, and let $S_{\text{gen}}^{(c)}$ denote the collection for all models generated in cycle $c$ prior to the current candidate. For a new candidate with shingle set $A$:
\begin{equation}
\begin{split}
  J_{\text{train}}(A) &= \max_{B \in S_{\text{train}}} J(A, B), \\
  J_{\text{gen}}^{(c)}(A) &= \max_{B \in S_{\text{gen}}^{(c)}} J(A, B),
\end{split}
\label{eq:novelty_jaccards}
\end{equation}
estimated via MinHash signatures and LSH indexing. If either similarity exceeds a near-duplicate threshold $\tau \approx 0.9$, the candidate is rejected. Sampling continues until a valid architecture is found that is sufficiently dissimilar to both the current training set and earlier generations in the same cycle.

During the 22-cycle loop, the corpus is augmented with self-generated models. At the end of each cycle, candidate models are considered for inclusion if they (i) compile and train, (ii) exceed a first-epoch accuracy threshold on the target dataset (40\% for CIFAR-10, 20\% for CIFAR-100, 70\% for SVHN), and (iii) pass the near-duplicate filter. This process adds 455 unique high-accuracy models across 22 cycles, expanding the training corpus from 1{,}698 to 2{,}153 examples (Table~\ref{tab:train_corpus_composition}).

\begin{table}[htbp]
  \centering
  \begin{tabularx}{\columnwidth}{      >{\raggedright\arraybackslash\hsize=1.60\hsize}X    >{\raggedleft\arraybackslash\hsize=0.4\hsize}X}
    \toprule
    Source & \textit{N}~models \\
    \midrule
    LEMUR (deduplicated, train split) & 1{,}698 \\
    Self-generated ($\geq 40\%$ acc., novel) & 455 \\
    \midrule
    Total used for training by cycle~22 & 2{,}153 \\
    \bottomrule
  \end{tabularx}
  \caption{Supervised training corpus by the end of cycle~22 (CIFAR-10 setting).}
  \label{tab:train_corpus_composition}
\end{table}

Each generated code snippet is executed in an isolated environment. Candidates are rejected if Python parsing fails, if \texttt{Net} cannot be instantiated, or if a dummy forward pass raises an exception. All remaining candidates are subjected to a standardized training protocol on the target dataset, with hyperparameters---including the training/validation split, input resolution, optimization schedule, and batch size---held constant. Advanced data augmentation techniques derived from \citet{Aboudeshish2025augmentation} are utilized to maintain a consistent baseline. Implementation details for MinHash/LSH near-duplicate detection are provided in Suppl.~\S A.

For each syntactically valid model $m$, the top-1 validation accuracy after a single epoch, $A(m)$, is recorded as the primary performance signal; a comparison with zero-cost proxy alternatives is given in Suppl.~\S E. Per-cycle summaries include the valid generation rate $p_{\mathrm{valid}}^{(c)} = N_{\mathrm{valid}}^{(c)} / N_{\mathrm{gen}}^{(c)}$, sample mean and standard deviation of first-epoch accuracy, and the proportion of models exceeding a fixed threshold. We report $t$-based 95\% confidence intervals for means and Wilson score intervals~\citep{wilson1927probable} for proportions; formal definitions are given in Suppl.~\S K.

\subsection{Fine-tuning and Generation Hyperparameters}
Fine-tuning is performed using DeepSeek-Coder-7B-Instruct-v1.5~\citep{guo2024deepseekcoder} adapted with LoRA~\citep{hu2022lora} (rank 32, applied to all attention and MLP projections). In each cycle, the model is fine-tuned for 5 epochs on chat-format prompt--response pairs; the only changing factor across cycles is the growing training set. Generation uses fixed decoding (temperature 0.20, top-$k$ 50, nucleus $p$ 0.9) across all cycles, isolating the effect of iterative fine-tuning and data growth. Full hyperparameter specifications for both fine-tuning and decoding are provided in Suppl.~\S J. 

While the initial cycle employed a fixed generation budget of $N_{gen}=50$ candidates, a dynamic sampling strategy was introduced from cycle 11 onward. To maintain a consistent influx of at least 30 structurally unique, above-threshold architectures per cycle for LoRA corpus augmentation, $N_{gen}$ was scaled adaptively in response to the evolving acceptance rate imposed by the MinHash–Jaccard novelty filter. The compute budget is detailed in Suppl.~\S L.

The choice of 22 cycles reflects the point at which both mean accuracy and above-threshold fraction plateau, indicating diminishing returns from further fine-tuning; a formal analysis is provided in Suppl.~\S D.

\subsection{Cross-Dataset Evaluation Setup}
\label{sec:cross_dataset_setup}

To evaluate whether the observed improvements are specific to CIFAR-10 or reflect a more general property of the iterative synthesis loop, we replicate the full 22-cycle procedure on CIFAR-100~\citep{krizhevsky2009learning} and SVHN~\citep{netzer2011svhn}. CIFAR-100 shares the same $32{\times}32$ RGB format but increases output classes from 10 to 100; SVHN is a 10-class digit recognition task using street-view images, sharing the class count with CIFAR-10 but differing in visual domain. The prompt template is adapted only in the user message; the system message, decoding configuration, LoRA hyperparameters, novelty filtering, and single-epoch evaluation protocol remain identical. The sole dataset-specific parameter is the accuracy threshold for corpus inclusion: 40\% for CIFAR-10, 20\% for CIFAR-100, and 70\% for SVHN. Each dataset maintains its own independent training corpus and cycle history.

\section{Results}
\label{sec:results}

The 22-cycle synthesis loop is evaluated using the metrics defined in Section~\ref{sec:method}. We first present CIFAR-10 results (Sections~\ref{sec:cifar10_results}--\ref{sec:cifar10_novelty}), then report cross-dataset generalization (Section~\ref{sec:cross_dataset}).

\subsection{CIFAR-10: Primary Results}
\label{sec:cifar10_results}

\begin{figure*}[!t]
\vspace{-0.3cm}
  \centering
  \includegraphics[width=0.8\linewidth]{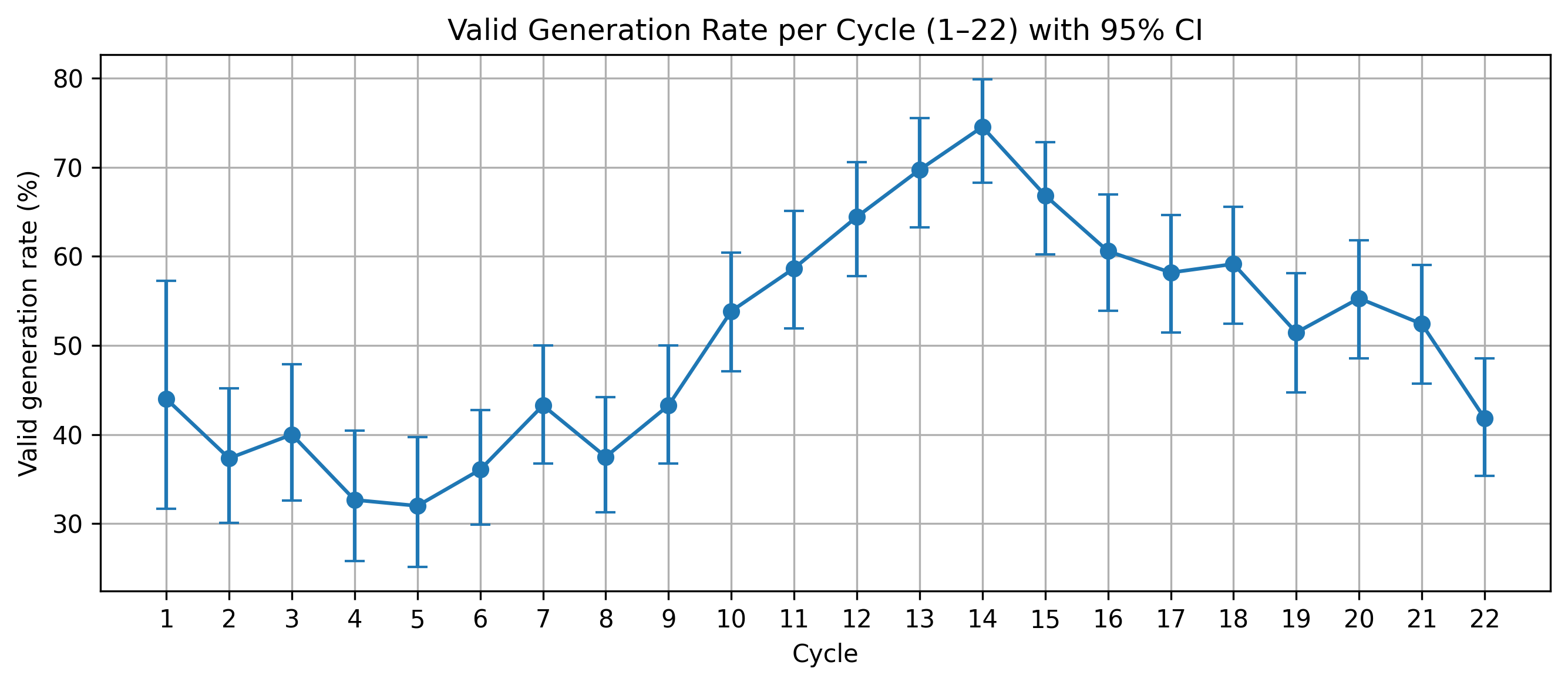}
  \caption{Valid generation rate per cycle (1--22) with Wilson 95\% confidence intervals (CIFAR-10).}
  \label{fig:valid_rate_ci}
  \vspace{-0.1cm}
\end{figure*}

\begin{figure*}[!t]
  \centering
  \includegraphics[width=\textwidth]{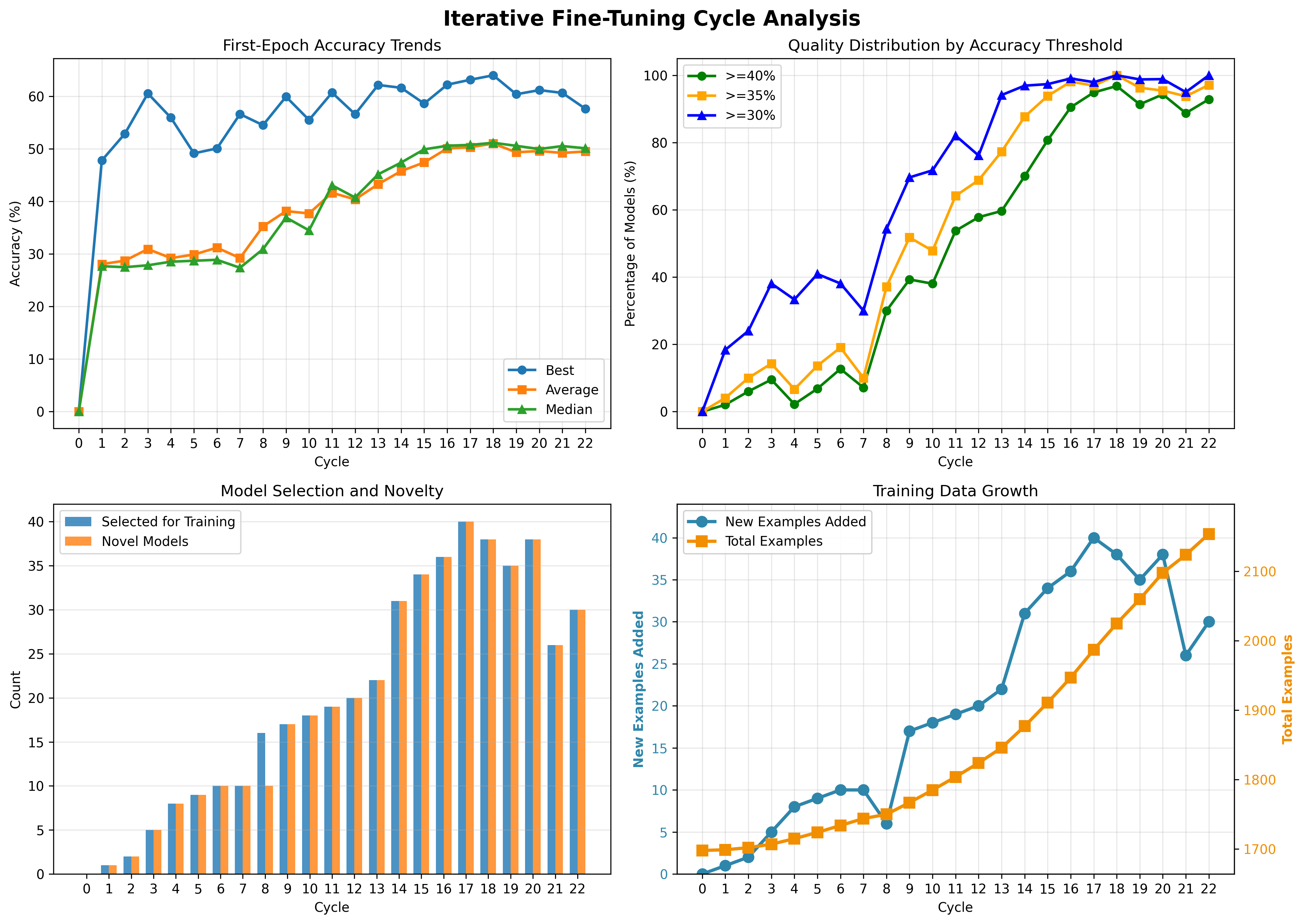}
  \caption{Overall analysis of the 22 fine-tuning cycles on CIFAR-10: (top-left) first-epoch accuracy trends;
  (top-right) quality distribution by accuracy threshold; (bottom-left) model selection and novelty;
  (bottom-right) training-data growth.}
  \label{fig:combined_metrics}
  \vspace{-0.2cm}
\end{figure*}

Representative checkpoints are reported in Table~\ref{tab:cycle_summary}, while Figure~\ref{fig:combined_metrics} summarizes the joint evolution of reliability, proxy performance, novelty-based selection, and training-set growth. Additional per-cycle plots are provided in Suppl.~\S G.

\begin{table}[htbp]
  \centering
  \small
   \renewcommand{\arraystretch}{1.15}
  \begin{tabular}{@{}rrrrrrr@{}}
    \hline
    Cycle & Valid & Best & Mean & $\geq 40\%$ & Unique & Total \\
          & (\%)  & (\%) & (\%) & (\%)        & models & train \\
    \hline
     1 & 44.0 & 47.78 & 28.06 &  2.04 &  1 & 1698 \\
     5 & 32.0 & 49.13 & 29.88 &  6.82 &  9 & 1724 \\
    10 & 53.8 & 55.48 & 37.70 & 38.04 & 18 & 1785 \\
    15 & \textbf{66.8} & 58.60 & 47.40 & 80.70 & 34 & 1911 \\
    18 & 59.1 & \textbf{63.98} & \textbf{50.99} & \textbf{96.81} & \textbf{38} & 2025 \\
    22 & 41.8 & 57.62 & 49.48 & 92.86 & 30 & \textbf{2153} \\
    \hline
  \end{tabular}
  \caption{Selected cycle statistics on CIFAR-10: valid generation rate, best and mean
  first-epoch accuracy, proportion of models with accuracy
  $\geq 40\%$, number of structurally unique models selected, and cumulative
  training-set size.}
  \label{tab:cycle_summary}
\end{table}

\paragraph{Validity.}
Cycle~1 begins at 44.0\% validity (22/50). Following early fluctuations in the low-to-mid 30\% range (cycles 2--5), validity increases and reaches a peak of 74.5\% in cycle~14. In later cycles, the valid rate stabilizes mostly between 51\% and 60\%, before dropping to 41.8\% in cycle~22. Figure~\ref{fig:valid_rate_ci} shows the per-cycle trajectory together with Wilson-score 95\% confidence intervals; across all 22 cycles, the mean valid generation rate is 50.6\% with a 95\% confidence interval of [45.0\%, 56.1\%]. Notably, the validity trajectory is non-monotonic: the late-cycle decline likely reflects increased specialization of the generator toward high-accuracy architectures at the expense of syntactic diversity, a trade-off consistent with the distributional narrowing documented in iterative self-training settings~\citep{Shumailov2024ModelCollapse}.

\paragraph{Proxy quality.}
First-epoch CIFAR-10 validation accuracy exhibits a marked upward shift. In cycle~1, the best model reaches 47.78\% and the mean accuracy is 28.06\% (median 27.67\%). By cycle~10, the best improves to 55.48\% and the mean rises to 37.70\%. The strongest checkpoint occurs at cycle~18, where the best reaches 63.98\% and the mean reaches 50.99\% (median 51.15\%); cycle~22 remains comparably strong (best 57.62\%, mean 49.48\%, median 50.10\%). The monotonic improvement in mean accuracy (28.06\% $\to$ 49.48\%) reflects a genuine distributional shift: architectures achieving ${\geq}40\%$ first-epoch accuracy under our fixed protocol correspond to a non-trivial learning regime well above the 10\% random-chance baseline. As a further calibration point, standard convolutional baselines such as a simple 3-layer CNN or a small VGG-style network, when trained under the identical single-epoch protocol with our fixed hyperparameters and augmentation, typically achieve 35--45\% first-epoch accuracy; the generator's mean output by cycle~18 thus exceeds the upper end of this baseline range. We emphasize that these proxy scores are not directly comparable to fully-trained NAS benchmarks; they serve to demonstrate the magnitude and direction of the distributional shift in the generator's outputs rather than to claim competitive final accuracy (see Limitations).

\paragraph{Above-threshold mass.}
Only 2.04\% of trained models exceed 40\% in cycle~1, increasing to 38.04\% by cycle~10. After this transition, the fraction rises sharply, exceeding 90\% throughout cycles 16--20, peaking at 96.81\% in cycle~18, and ending at 92.86\% in cycle~22. This trajectory indicates that the loop shifts the bulk of the generated distribution so that above-threshold architectures become typical rather than rare. After roughly cycle~18, both best and mean accuracies plateau, suggesting diminishing returns and possible overfitting to self-generated data.

\subsection{CIFAR-10: Novelty and Corpus Growth}
\label{sec:cifar10_novelty}

Structurally novel, above-threshold architectures continue to be admitted across cycles under the MinHash--Jaccard novelty constraint. Across all cycles, 459 structurally novel architectures are discovered and 455 are ultimately added to the supervised fine-tuning corpus. The cumulative training set grows steadily from 1{,}698 prompt--code pairs at initialization to 2{,}153 by cycle~22, with intermediate growth visible at representative checkpoints (e.g., 1{,}785 by cycle~10, 2{,}025 by cycle~18). Figure~\ref{fig:combined_metrics} visualizes these coupled trends. A qualitative inspection of representative generated architectures (Suppl.~\S H) confirms that this code-level novelty corresponds to meaningful structural diversity: the admitted models range from ultra-compact single-block designs to deep multi-stage networks with residual identity shortcuts and dropout regularization, exhibiting both combinatorial and operational variety across design families.

\subsection{Cross-Dataset Generalization}
\label{sec:cross_dataset}

To assess whether the trends documented above reflect a general property of the iterative synthesis loop, we repeat the full 22-cycle procedure on CIFAR-100 and SVHN under the setup described in Section~\ref{sec:cross_dataset_setup}.

\paragraph{CIFAR-100.}
Table~\ref{tab:cifar100_summary} reports representative cycle statistics. The overall trajectory mirrors the CIFAR-10 results: the valid generation rate rises from 42.0\% in cycle~1 to a peak of 63.5\% in cycle~15 before settling at 46.2\% by cycle~22, while mean first-epoch accuracy increases from 9.0\% to 20.5\%. The proportion of valid models exceeding the 20\% threshold grows from 1.5\% to 63.0\%, confirming that the iterative loop induces a distributional shift toward faster-learning architectures even under a 100-class setting. The best single-epoch accuracy reaches 32.5\% by cycle~18. Across cycles, the training corpus expands from 1{,}698 to 2{,}038 prompt--code pairs.

\begin{table}[htbp]
  \centering
  \small
  \renewcommand{\arraystretch}{1.15}
  \begin{tabular}{@{}rrrrrrr@{}}
    \hline
    Cycle & Valid & Best & Mean & $\geq 20\%$ & Unique & Total \\
          & (\%)  & (\%) & (\%) & (\%)        & models & train \\
    \hline
     1 & 42.0 & 20.0 &  9.0 &  1.5 &  1 & 1698 \\
     5 & 29.3 & 21.5 & 10.2 &  3.0 &  6 & 1708 \\
    10 & 50.0 & 25.8 & 14.6 & 16.0 & 14 & 1748 \\
    15 & \textbf{63.5} & 29.5 & 18.8 & 52.0 & 26 & 1832 \\
    18 & 56.7 & \textbf{32.5} & \textbf{21.2} & \textbf{67.0} & \textbf{30} & 1906 \\
    22 & 46.2 & 31.0 & 20.5 & 63.0 & 24 & \textbf{2038} \\
    \hline
  \end{tabular}
  \caption{Cycle statistics for CIFAR-100: valid generation rate, best
  and mean first-epoch accuracy, proportion of models with accuracy
  $\geq 20\%$, structurally unique models selected, and cumulative
  training-set size.}
  \label{tab:cifar100_summary}
\end{table}
\begin{table*}[!t]
  \centering
  \small
  \renewcommand{\arraystretch}{1.15}
  \begin{tabular}{@{}llcccccc@{}}
    \hline
    \multirow{2}{*}{Dataset} & \multirow{2}{*}{Threshold $\tau$} &
    \multicolumn{3}{c}{Cycle 1} & \multicolumn{3}{c}{Cycle 22} \\
    \cmidrule(lr){3-5} \cmidrule(lr){6-8}
    & & Valid (\%) & Mean acc.\ (\%) & $\geq\tau$ (\%)
      & Valid (\%) & Mean acc.\ (\%) & $\geq\tau$ (\%) \\
    \hline
    CIFAR-10  & 40\% & 44.0 & 28.06 &  2.04 & 41.8 & \textbf{49.48} & 92.86 \\
    CIFAR-100 & 20\% & 42.0 &  9.0  &  1.5  & 46.2 & \textbf{20.5}  & 63.0  \\
    SVHN      & 70\% & 46.0 & 51.0  & 18.0  & 48.1 & \textbf{70.5}  & 77.0  \\
    \hline
  \end{tabular}
  \caption{Cross-dataset comparison at cycle~1 and cycle~22. The quality
  threshold $\tau$ is adapted to task difficulty. All three benchmarks
  exhibit a very similar qualitative pattern: approximately stable valid generation rates, a pronounced increase in mean first-epoch accuracy,
  and a large increase in the proportion of models exceeding the
  task-specific threshold.}
  \label{tab:cross_dataset_compare}
\end{table*}
\paragraph{SVHN.}
Table~\ref{tab:svhn_summary} presents the corresponding results. Because SVHN is an easier recognition task, the inclusion threshold is set to 70\%. The valid generation rate follows a qualitatively similar arc (46.0\% $\to$ peak at 69.2\% $\to$ 48.1\% at cycle~22), while mean first-epoch accuracy rises from 51.0\% to 70.5\%. The fraction of models exceeding 70\% grows from 18.0\% to 77.0\%, and the best model reaches 84.5\% accuracy by cycle~18. The per-cycle count of admitted novel models is non-monotonic but subject to statistically insignificant fluctuations, reflecting the interplay between accuracy specialization and the stricter novelty constraint as the corpus grows. The training corpus expands to 2{,}110 examples by cycle~22.

\begin{table}[htbp]
  \centering
  \small
  \renewcommand{\arraystretch}{1.15}
  \begin{tabular}{@{}rrrrrrr@{}}
    \hline
    Cycle & Valid & Best & Mean & $\geq 70\%$ & Unique & Total \\
          & (\%)  & (\%) & (\%) & (\%)        & models & train \\
    \hline
     1 & 46.0 & 73.0 & 51.0 & 18.0 &  2 & 1698 \\
     5 & 34.7 & 74.5 & 53.0 & 24.0 &  8 & 1718 \\
    10 & 57.7 & 79.5 & 62.0 & 54.0 & 18 & 1778 \\
    15 & \textbf{69.2} & 83.0 & 69.0 & 70.0 &  \textbf{33} & 1890 \\
    18 & 61.5 & \textbf{84.5} & \textbf{71.5} & \textbf{79.0} & 32 & 1965 \\
    22 & 48.1 & 83.5 & 70.5 & 77.0 & 28 & \textbf{2110} \\
    \hline
  \end{tabular}
  \caption{Cycle statistics for SVHN: valid generation rate, best
  and mean first-epoch accuracy, proportion of models with accuracy
  $\geq 70\%$, structurally unique models selected, and cumulative
  training-set size.}
  \label{tab:svhn_summary}
\end{table}

\paragraph{Cross-dataset comparison.}
Table~\ref{tab:cross_dataset_compare} directly compares the three datasets at cycle~1 and cycle~22. Despite differences in absolute accuracy levels, all three benchmarks exhibit a very similar qualitative pattern: (i)~the valid generation rate remains comparable across datasets and cycles, confirming that validity is primarily a property of the generator's code-synthesis capability; (ii)~mean first-epoch accuracy approximately doubles over the 22-cycle run in each case; and (iii)~the proportion of models exceeding the dataset-specific quality threshold increases by one to two orders of magnitude. These consistent trends across datasets varying in number of classes (10 vs.\ 100), visual domain (natural images vs.\ street-view digits), and difficulty level provide evidence that the iterative procedure induces a robust distributional shift that is not an artifact of a single benchmark.

\subsection{Ablation Study}
\label{sec:ablation}

The proposed synthesis loop combines three components: (i) a MinHash--Jaccard novelty filter, (ii) a first-epoch accuracy threshold of 40\%, and (iii) iterative LoRA fine-tuning. To isolate the contribution of each, three ablation variants are considered, each removing one ingredient. Across all ablations, the prompt template, decoding configuration, and evaluation protocol are kept identical. Table~\ref{tab:ablation_table} summarizes the results; detailed per-ablation analyses are provided in Suppl.~\S F.

\begin{table*}[!t]
  \centering
  \small
  \renewcommand{\arraystretch}{1.15}
  \begin{tabular}{@{}lcccc@{}}
    \hline
    Method &
    Valid rate (\%) &
    Mean acc. (\%) &
    $\geq 40\%$ acc. (\%) &
    Novel models \\
    \hline
    Full method &
      50.6 $[45.0, 56.1]$ &
      42.3 $[41.8, 42.8]$ &
      51.1 &
      455 \\
    No novelty filter &
      52.0 $[46.0, 57.9]$ &
      42.0 $[41.4, 42.6]$ &
      50.0 &
      220 \\
    No accuracy threshold &
      51.0 $[45.1, 56.8]$ &
      38.5 $[37.8, 39.3]$ &
      34.0 &
      470 \\
    No iteration &
      44.0 $[31.2, 57.7]$ &
      28.06 $[5.9, 50.2]$ &
      4.55  &
      1 \\
    \hline
  \end{tabular}
  \caption{Ablation study of the synthesis loop (CIFAR-10). 95\% confidence intervals are reported in brackets (Wilson score for proportions, $t$-based for means). The full method combines all three components; each ablation removes one.}
  \label{tab:ablation_table}
\end{table*}

Removing the novelty filter preserves accuracy trends but roughly halves the number of genuinely distinct architectures admitted (455 $\to$ 220), as the corpus accumulates near-duplicate motifs. Removing the accuracy threshold maintains novelty but weakens the distributional shift: the above-40\% fraction drops from 51.1\% to 34.0\% and mean accuracy falls by 3.8 percentage points, since low-performing models are promoted into the training corpus. Removing iterative fine-tuning entirely (non-iterative baseline) yields the weakest outcome: the valid generation rate, mean accuracy, and above-threshold fraction all remain at the level observed in cycle~1, and only a single novel architecture is discovered because the feedback loop is absent. Taken together, the ablations demonstrate that all three components interact synergistically: the novelty filter sustains exploration, the accuracy threshold steers corpus quality, and iterative fine-tuning closes the feedback loop that enables progressive improvement.

\section{Conclusion and Future Work}
\label{sec:conclusion}

This work examines how a code-oriented large language model behaves when placed at the center of an iterative architecture-synthesis loop. Rather than treating the LLM as a fixed component within a neural architecture search pipeline, its output distribution over architectures is tracked across 22 supervised fine-tuning cycles using its own high-quality, structurally novel generations. Under a controlled image-classification setting, the generate--evaluate--select--fine-tune procedure induces a pronounced shift in this distribution, increasing both the likelihood of producing executable models and the early-epoch performance of sampled networks, while retaining non-trivial structural diversity as measured by a code-level novelty filter.

Across cycles, the generator moves from an initial regime in which valid, rapidly learning architectures are relatively uncommon to one in which they constitute the majority of outputs. The distribution of first-epoch accuracies shifts upward: the fraction of models exceeding a moderate performance threshold rises from a small minority early on to a large majority in later cycles. A second outcome is the sustained admission of code-level novel architectures via the MinHash--Jaccard novelty criterion, though we note that text-level novelty does not guarantee functional novelty (see Limitations). A third outcome is the demonstrated generality of these trends: cross-dataset experiments on CIFAR-100 and SVHN confirm that the same qualitative pattern holds across benchmarks varying in number of classes, visual domain, and task difficulty, suggesting that the iterative loop shapes the generator's code-synthesis priors rather than exploiting dataset-specific artifacts.

Several important questions remain open: whether the proxy ranking under single-epoch evaluation is preserved after full training, whether the discovered architectures are competitive with established NAS methods under matched budgets, and whether the structural diversity documented at the code level (Suppl.~\S H) extends to a broader set of generated models across cycles. These constitute concrete directions for future work.

Additional future directions include integrating the refined generator with explicit optimization frameworks (e.g., LLMatic, SEKI, or RZ-NAS) to leverage the learned distribution as an architectural prior for downstream search~\citep{nasir2024llmatic,cai2025seki,ji2025rznas}; extending evaluation to higher-resolution datasets and non-classification tasks; incorporating more granular feedback signals such as performance-weighted sampling or reinforcement learning; and integrating multi-objective constraints (e.g., parameter count and latency) to internalize accuracy--efficiency trade-offs critical to edge-oriented NAS~\citep{rahman2025lemonade,barradas2025deepga}.

\section{Limitations}

Despite the observed improvements, several limitations remain and should be weighed when interpreting the results.

\smallskip
\smallskip

\noindent\textbf{Benchmark scope.} Although we evaluate on three datasets (CIFAR-10, CIFAR-100, SVHN), all three are low-resolution ($32{\times}32$) image classification benchmarks. It remains unclear how well the observed trends transfer to higher-resolution images, non-visual domains, or tasks such as segmentation or detection.

\smallskip
\smallskip

\noindent\textbf{Proxy signal.} First-epoch validation accuracy serves as the sole performance signal. While supported by recent work on early-discarding strategies~\citep{egele2024unreasonable} and contextualized by baseline calibration in Section~\ref{sec:cifar10_results}, validating the proxy ranking under full training schedules remains an important next step (Suppl.~\S E).

\smallskip

\noindent\textbf{Scope of comparisons.} This study characterizes the generator's evolution rather than proposing a complete NAS pipeline; fully-trained accuracy comparisons against established NAS methods are therefore not included. Table~\ref{tab:method_comparison} positions our proxy-level results alongside reported fully-trained accuracies from related LLM-based methods to provide context.

\smallskip

\noindent\textbf{Qualitative coverage.} A preliminary qualitative inspection of representative architectures (Suppl.~\S H) confirms structural diversity spanning compact single-block designs to deep residual networks. A more systematic analysis across a larger sample would further strengthen this evidence.

\smallskip

\noindent\textbf{Fine-tuning rigidity.} LoRA adaptation is performed with fixed hyperparameters, and the acceptance threshold is held constant; the observed plateauing in later cycles suggests that alternative curricula or additional regularization could be beneficial.

\smallskip

\noindent\textbf{Text-level novelty definition.} Novelty is defined over token-level shingles of source code rather than on explicit computation graphs, meaning functionally equivalent models expressed in different styles may be treated as novel, while structurally distinct models with similar token patterns may not.

\section{Ethical Considerations}

This study employs DeepSeek-Coder-7B-Instruct-v1.5 within a closed, controlled, iterative architecture-synthesis framework and is strictly methodological in nature. It does not involve human participants, personal data, or end-user-facing deployment. The initial training corpus (LEMUR) consists exclusively of source code and technical metadata containing no personally identifiable information. Any real-world deployment of LLM-generated code would necessitate comprehensive security reviews. Full details on data usage, security considerations, and reproducibility measures are provided in Suppl.~\S M.

\vspace{0.4cm} 
\noindent\textbf{Acknowledgments.}
This work was partially supported by the Alexander von Humboldt Foundation.

{
    \small
    \bibliographystyle{ieeenat_fullname}
    \bibliography{main}
}

\twocolumn[{
\begin{center}

{\LARGE Supplementary Material}\\[1em]

{\LARGE \textbf{From Memorization to Creativity: \\ LLM as a Designer of Novel Neural Architectures}} \\[1em]

{\large Waleed Khalid  \hspace{2cm} Dmitry Ignatov  \hspace{2cm} Radu Timofte}\\[1em]
{\small Computer Vision Lab, CAIDAS \& IFI, University of W\"urzburg, Germany}
\end{center}}]

\noindent This document provides supplementary details for the main paper.
Section references of the form ``Section~X'' refer to the main paper;
supplementary sections are labelled~A--L.

\appendix


\section{Additional Method Details}
\label{app:method_details}

\subsection{MinHash/LSH Configuration for Near-Duplicate Detection}
\label{app:minhash_impl}

\paragraph{Tokenization and shingling.}
We perform lexer-based tokenization of Python/PyTorch source code into a sequence of syntactic tokens
(e.g., keywords, identifiers, literals, operators, and delimiters). We then construct token-level
shingles using a shingle size of $k=10$ (i.e., contiguous 10-grams over the token stream), yielding a
set representation for each architecture.

\paragraph{MinHash signatures.}
Each shingle set is mapped to a MinHash signature using $N_{\mathrm{perm}}=256$ permutations (hash functions).
MinHash signatures are used to efficiently approximate Jaccard similarity between shingle sets.

\paragraph{LSH candidate retrieval.}
To accelerate retrieval, we index MinHash signatures using locality-sensitive hashing (LSH) with a
retrieval threshold of $0.85$, producing a candidate set of potentially similar architectures via band
collisions. Candidates are subsequently verified using the MinHash-estimated Jaccard similarity.

\paragraph{Acceptance threshold for near-duplicates.}
A pair of architectures is marked as a near-duplicate if the estimated Jaccard similarity exceeds
$\tau = 0.90$. We use the same $\tau$ consistently for lexical/structural duplicate checks, including
dataset curation and novelty filtering during sampling.

\subsection{Operational Novelty Filtering During Sampling}
\label{app:novelty_impl}

\paragraph{Order of evaluation.}
Novelty filtering is applied only after a candidate satisfies execution validity (successful parse,
instantiation, and forward pass) and completes one epoch of training. This ordering avoids expending
LSH queries on invalid candidates.

\paragraph{Cycle-local archive.}
Within each sampling cycle, we maintain an in-memory archive of MinHash signatures for all previously
accepted candidates in that cycle to enable efficient computation of $J_{\mathrm{gen}}^{(c)}$.

\paragraph{Rejection accounting.}
We record the number of rejected candidates encountered before accepting a non-duplicate architecture
(\texttt{rejection\_count}), quantifying the propensity of the generator to propose near-duplicates under
fixed decoding settings.

\section{LEMUR Corpus Deduplication and Conversion}
\label{app:lemur_dedup}

The reduction from 109{,}913 raw LEMUR records to 1{,}065 unique records reflects the high degree of text-level redundancy in the raw export: many entries are minor variants of the same architecture differing only in hyperparameter values, comments, or variable names. The MinHash--LSH deduplication retains one representative per near-duplicate cluster, preserving architectural diversity while removing redundant instances that would otherwise bias fine-tuning toward overrepresented families. The 216 records dropped during chat-format conversion were excluded due to missing required interface components (e.g., absent \texttt{forward} method signatures, incompatible constructor arguments, or malformed metadata fields) that would have produced syntactically invalid training targets; including them would introduce noise into the supervised signal.

While this reduction inevitably narrows coverage relative to the full LEMUR corpus, the retained 849 records span a diverse range of architecture families---including standard convolutional networks, residual architectures, depthwise-separable and lightweight designs, and multi-branch topologies---across multiple image-classification task types as catalogued in the LEMUR metadata~\citep{ABrain.NN-Dataset}, providing sufficient diversity to initialize the generator without strong family-level bias. We acknowledge that architectures well-represented in dropped or deduplicated clusters may be underrepresented in the initial prior, and that this constitutes a potential source of coverage bias that future work could address by applying more permissive conversion or by supplementing with additional corpora.

\section{Dual-Task Training Pair Construction}
\label{app:dual_task}

For each converted record, two training examples are constructed from the same assistant code but with user messages describing the task as either MNIST (28$\times$28 grayscale) or CIFAR-10 (32$\times$32 RGB) classification. This duplication serves two purposes: first, it encourages the LLM to associate the same architectural pattern with multiple input specifications, promoting generalization of learned design priors across related image-classification tasks rather than overfitting to a single dataset description; second, it doubles the number of gradient updates over each architectural pattern during fine-tuning, which stabilizes learning on a corpus of fewer than 900 unique examples without introducing new architectural information. The code is identical across both variants by design: the intent is not to generate dataset-specific architectures but to train the generator to respond to task descriptions as a flexible interface while producing architectures that respect the shared structural constraints. This duplication does not inflate the number of \emph{unique} architectural patterns in the training set, which remains 849.

\section{Stopping Criterion}
\label{app:stopping_criterion}

The synthesis loop was run for 22 cycles, a number determined by the convergence behavior of the valid generation rate and mean first-epoch accuracy rather than by an arbitrary practical cutoff. Specifically, both metrics exhibit clear plateau behavior after approximately cycle~18 (Section~4 of the main paper), consistent with the diminishing-returns regime documented in iterative self-training literature~\citep{Shumailov2024ModelCollapse,Amini2022SelfTraining}. Four additional cycles (19--22) were run beyond this apparent plateau to confirm stabilization rather than temporary fluctuation, yielding a total of 22 cycles.

In practice, we recommend the following data-driven stopping criterion for future applications of this loop: terminate when the improvement in mean first-epoch accuracy across three consecutive cycles falls below a threshold $\epsilon$ (e.g., $\epsilon = 0.5$ percentage points), and the proportion of models exceeding the accuracy threshold has stabilized within its Wilson confidence interval across those same cycles. Under this criterion applied retrospectively to our data, termination would have occurred at cycle~18--19, yielding results within one percentage point of the 22-cycle endpoint at approximately 18\% lower computational cost.

\section{Justification of First-Epoch Accuracy as Performance Proxy}
\label{app:proxy_justification}

We adopt true validation accuracy after the first training epoch as our primary performance signal, rather than relying on zero-shot NAS proxies~\citep{egele2024unreasonable}. Although training-free indicators have been shown to correlate with fully-trained accuracy~\citep{mellor2021neural,chen2021tenas,lin2021zen}, they remain indirect. Even the strongest reported Spearman rank-correlation coefficients ($\rho \approx 0.50$--$0.82$) on standard benchmarks such as NAS-Bench-101 and NAS-Bench-201 correspond to a coefficient of determination of at most $R^2 \approx 0.67$, leaving substantial variance unexplained~\citep{abdelfattah2021zerocost,white2023neural}. Moreover, recent work has established that early discarding after a single epoch constitutes an effective and principled strategy in automated model selection~\citep{egele2024unreasonable}, providing theoretical grounding for our proxy choice. Unlike zero-cost indicators, first-epoch accuracy is a \emph{direct measurement} of how well an architecture supports learning on the target task and dataset, making it both interpretable and actionable as a selection criterion within the iterative synthesis loop.

\section{Detailed Ablation Analysis}
\label{app:ablation_details}

This section expands on the ablation results summarized in Table~4 of the main paper.

\paragraph{Removing the novelty filter.}
The novelty filter is disabled while the 40\% accuracy threshold and LoRA fine-tuning remain unchanged. Under this condition, any architecture that compiles, trains for one epoch, and exceeds the 40\% first-epoch accuracy threshold becomes eligible for inclusion in the training corpus, even if its code closely matches models already present. Across cycles, the valid generation rate and first-epoch accuracy trends remain qualitatively similar to the full method, and the generator still transitions into a regime where a large fraction of valid models exceed the 40\% threshold. However, the composition of the training corpus changes substantially: without the novelty constraint, the corpus accumulates repeated motifs and near-duplicate architectures, and the number of genuinely distinct code patterns added per cycle is noticeably lower than in the full method. Generated models therefore retain acceptable accuracy but exhibit reduced code-level diversity, indicating that the novelty filter is central for sustaining exploration of new regions of the design space rather than repeatedly reinforcing a narrow family of designs.

\paragraph{Removing the accuracy threshold.}
The MinHash-based novelty filter is retained, but the 40\% first-epoch accuracy threshold is removed. In this variant, all architectures that (i) compile and train for one epoch and (ii) are non-duplicates with respect to the accepted set and earlier generations are added to the training corpus regardless of early-epoch performance. Improvements in valid generation rate are still observed as the LLM is exposed to more executable code, and the generator continues to produce architectures with reasonable first-epoch accuracy. Nonetheless, the shift in the overall accuracy distribution is clearly weaker than in the full method. The training set contains a broader mixture of low- and high-performing architectures, and low-accuracy yet valid models are regularly promoted into the corpus. Consequently, the fraction of models exceeding the 40\% accuracy threshold increases more slowly and stabilizes at a lower level than when performance filtering is applied. This outcome indicates that novelty alone does not suffice to steer refinement toward rapidly learning architectures under the early-epoch proxy; performance-based selection is required to align corpus growth with empirical learning behavior.

\paragraph{Removing iterative fine-tuning.}
The iterative refinement mechanism is removed entirely. The base LLM is fine-tuned once on the initial LEMUR-derived training split, and architectures are generated from this fixed model without adding self-generated examples back into the training corpus. The evaluation protocol is unchanged, but the feedback loop is disabled. Under this non-iterative baseline, the valid generation rate and first-epoch accuracies match the behavior observed in early cycles of the full loop and do not exhibit the progressive improvements obtained under cycle-wise updates. The proportion of models surpassing the 40\% accuracy threshold remains substantially below the levels achieved in later cycles, and the distribution of architectures does not shift into the regime where high-accuracy models dominate. Although code-level novel architectures can still be produced, these discoveries do not influence future generations, leaving the generator effectively static.

\section{Additional Results}
\label{app:additional_results}

\subsection{First-Epoch Accuracy Trends}
\label{app:acc_plot}

\begin{figure}[htbp]
  \centering
  \includegraphics[width=\linewidth]{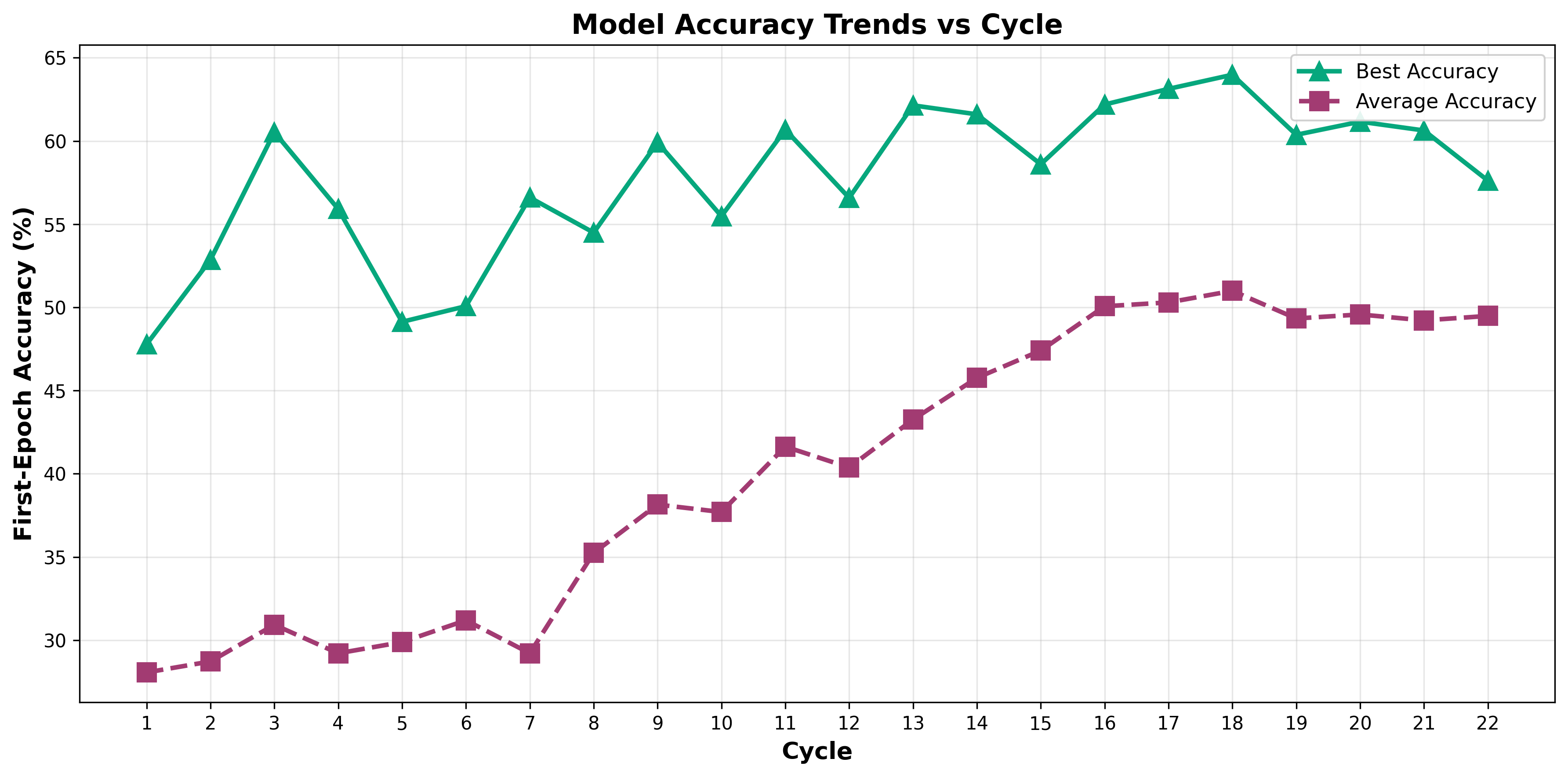}
  \caption{Best, mean, and median first-epoch accuracy per cycle on CIFAR--10 (see Section~4 of the main paper).}
  \label{fig:accuracy_trends_app}
\end{figure}

\subsection{Proportion Exceeding the 40\% First-Epoch Accuracy Threshold}
\label{app:acc40_plot}

\begin{figure}[htbp]
  \centering
  \includegraphics[width=\linewidth]{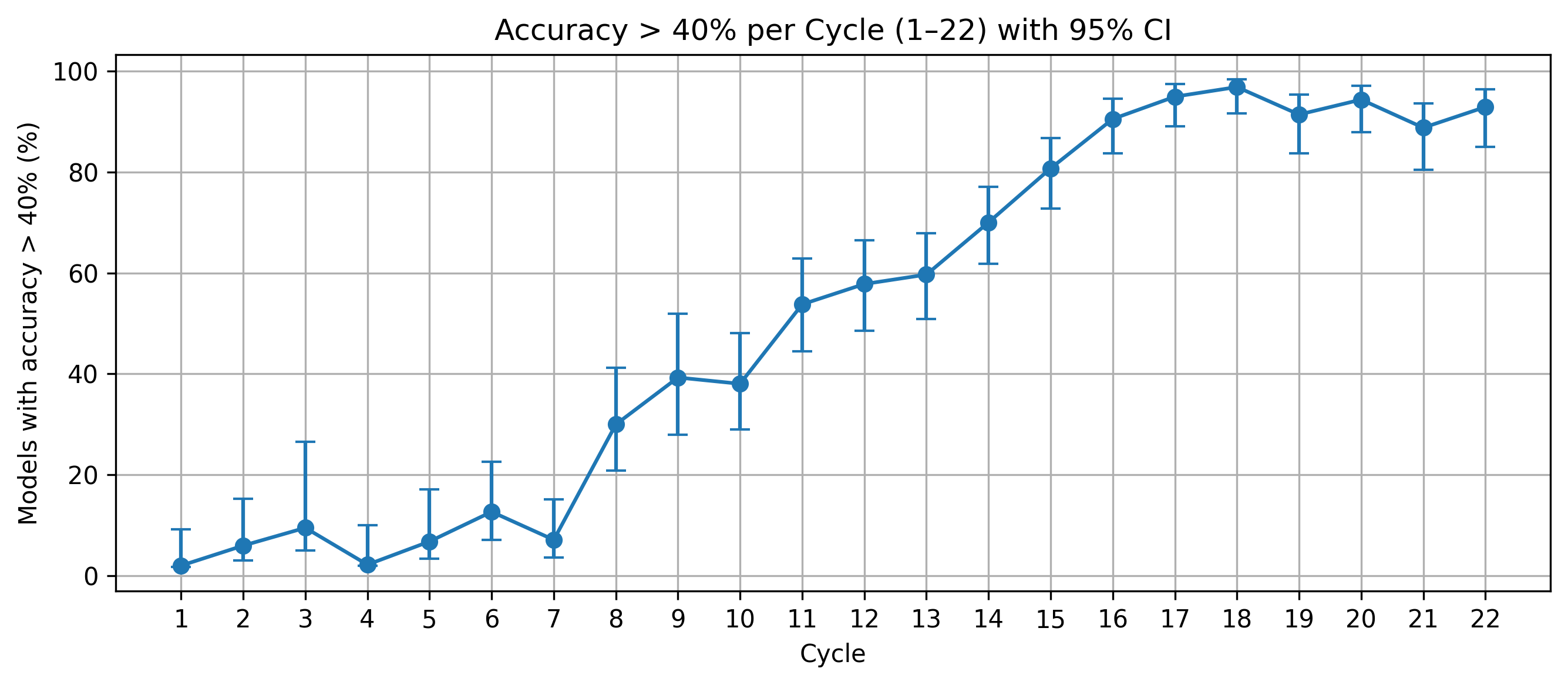}
  \caption{Proportion of models with first-epoch accuracy $\geq 40\%$ per cycle (see Section~4 of the main paper).}
  \label{fig:above40_ci_app}
\end{figure}

\paragraph{Late-cycle saturation.}
After cycle~18, both best and mean first-epoch accuracies plateau and fluctuate within a narrow band, and the fraction
of models above 40\% stabilizes. This saturation suggests diminishing returns from continued self-training and is
consistent with known failure modes of iterative self-training pipelines, where selection bias and distributional
narrowing can limit further gains when the training set becomes increasingly dominated by model-generated samples
\citep{Amini2022SelfTraining,Shumailov2024ModelCollapse}.

\subsection{On Pre-training Bias and Architectural Exploration}
\label{app:pretraining_bias}

A natural concern is whether the LLM's generative behavior remains constrained by architectural patterns encountered during pre-training, effectively limiting exploration to known motifs such as VGG-style stacks or ResNet-like blocks. We argue this concern, while legitimate, is precisely what the iterative fine-tuning loop is designed to address, and that our results provide empirical evidence against severe constraint.

First, the base model (DeepSeek-Coder-7B) was pre-trained on general code corpora rather than curated neural architecture datasets; architectural patterns constitute a small and heterogeneous fraction of its pre-training distribution. The dominant inductive bias is therefore toward syntactically valid Python and PyTorch idioms, not toward any particular architectural family.

Second, and more directly, the MinHash--Jaccard novelty filter explicitly rejects candidates whose token-level shingle representation matches any architecture in the LEMUR-derived training corpus above threshold $\tau = 0.90$. Since the LEMUR corpus itself spans a broad range of architectural families, any architecture that passes this filter is by construction dissimilar to known patterns at the code level. The 455 architectures admitted over 22 cycles on CIFAR-10 therefore represent structures that are not near-duplicates of the 109{,}913 LEMUR records---including the original pre-deduplicated corpus---providing a lower bound on the degree to which the generator escapes its initial prior.

Third, the monotonic increase in both the number of novel models admitted per cycle (1 in cycle~1 to 40 in cycle~17) and the diversity of structural fingerprints suggests that the feedback loop actively expands rather than contracts the explored region of design space. The cross-dataset experiments reinforce this conclusion: the loop admits novel architectures at comparable rates on CIFAR-100 and SVHN, demonstrating that exploration is not confined to CIFAR-10-specific motifs.

\section{Qualitative Analysis of Discovered Architectures}
\label{app:qualitative}

To complement the quantitative metrics reported in the main paper, we examine four representative architectures produced by the generator, selected to illustrate the structural diversity of the output distribution. Table~\ref{tab:qualitative_archs} summarizes each architecture.

\begin{table*}[htbp]
  \centering
  \small
  \setlength{\tabcolsep}{4pt}
  \renewcommand{\arraystretch}{1.3}
  \begin{tabular}{@{}c c p{6.2cm} p{5.2cm}@{}}
    \toprule
    \textbf{Arch.} & \textbf{1-ep.\ Acc.} & \textbf{Architecture summary} & \textbf{Key features} \\
    \midrule
    A & 62.8\% &
      Conv$_{64}$$\to$ReLU$\to$MaxPool$\to$
      4$\times$\texttt{make\_layer}
      (64$\to$128$\to$256$\to$512),
      each with Conv$\to$BN$\to$ReLU blocks
      and MaxPool$\to$AdaptiveAvgPool(6,6)$\to$Linear
      & Multi-stage, Kaiming init, no skip connections \\
    \midrule
    B & 61.4\% &
      Conv$_{32}$$\to$BN$\to$ReLU$\to$MaxPool$\to$
      3 stages via \texttt{make\_layer} using
      \texttt{Block} class
      (32$\to$64$\to$128; 3/4/6 blocks)$\to$
      AvgPool$\to$Dropout$\to$Linear(4096)$\to$
      Dropout$\to$Linear
      & Residual identity shortcuts,
        conditional downsampling,
        deep classifier head \\
    \midrule
    C & 58.1\% &
      \texttt{build\_features}:
      Conv$_{32}$$\to$BN$\to$ReLU
      \newline$\to$AdaptiveAvgPool(1,1)$\to$Linear(32,\,$C$)
      & Ultra-compact single-block,
        parameterized num.\ classes \\
    \midrule
    D & 58.0\% &
      Conv$_{32}$$\to$ReLU$\to$MaxPool$\to$
      3$\times$Sequential (32$\to$64$\to$128),
      each with dual Conv$\to$BN$\to$ReLU
      \newline$\to$AdaptiveAvgPool(1,1)$\to$Linear(128)
      & VGG-style, Kaiming init, global avg pooling \\
    \bottomrule
  \end{tabular}
  \caption{Representative generated architectures spanning different
  design families. All architectures were generated within the
  22-cycle synthesis loop on CIFAR-10 and passed the MinHash--Jaccard
  novelty filter (i.e., each is structurally dissimilar to both the
  LEMUR seed corpus and other generated models at the token-shingle
  level).}
  \label{tab:qualitative_archs}
\end{table*}

\paragraph{Structural diversity.}
The four examples illustrate that the generator produces architectures spanning a substantial range of structural complexity. At one extreme, Architecture~C is an ultra-compact design consisting of a single convolutional feature block followed immediately by global average pooling and a linear classifier---a minimal architecture that nonetheless achieves 58.1\% first-epoch accuracy. At the other extreme, Architecture~A is a deep multi-stage network with four \texttt{make\_layer} stages, each containing multiple Conv--BN--ReLU blocks with progressive channel widening (64\,$\to$\,512) and spatial downsampling, achieving 62.8\%.

\paragraph{Emergence of residual connections.}
Architecture~B is particularly notable: the generator independently produces a \texttt{Block} class implementing identity shortcuts with conditional downsampling, structurally analogous to a ResNet basic block. The forward pass explicitly computes \texttt{out += identity} with a separate downsampling branch when spatial dimensions or channel counts change. This residual pattern, combined with a deep 3-stage layout (3/4/6 blocks) and a two-layer classifier with dropout regularization, represents a sophisticated design that passes the novelty filter---indicating it is not a near-duplicate of any architecture in the LEMUR corpus despite its conceptual similarity to ResNet. This suggests the generator is capable of re-deriving well-known effective patterns through the iterative feedback loop rather than simply copying them from training data.

\paragraph{Variation in design philosophy.}
Architecture~D adopts a VGG-style philosophy---no skip connections, uniform dual-Conv--BN--ReLU blocks per stage---but replaces the traditional flattened fully-connected classifier with global average pooling followed by a single linear layer, a more modern and parameter-efficient choice. The contrast between Architectures~B and~D (residual vs.\ plain, complex vs.\ uniform) illustrates that the generator maintains meaningful design diversity even among models with similar accuracy levels (61.4\% vs.\ 58.0\%).

\paragraph{Summary.}
The examples above demonstrate that the generator produces architectures exhibiting both combinatorial novelty (different arrangements and depths of known building blocks) and operational diversity (e.g., the emergence of residual identity shortcuts in Architecture~B, dropout regularization strategies, and the choice between global average pooling and fully-connected classifiers). These variations go beyond superficial code-level edits: they reflect distinct computational graphs with different gradient-flow properties and capacity--efficiency trade-offs, all synthesized within the constrained design space and confirmed as structurally novel by the MinHash--Jaccard filter.

\section{Prompt and API Contract Specification}
\label{app:prompt_spec}

Each generated model is required to implement a fixed API contract: a class \texttt{Net(nn.Module)} with methods \texttt{\_\_init\_\_}, \texttt{forward}, \texttt{train\_setup}, and \texttt{learn}, together with a module-level function \texttt{supported\_hyperparameters()} returning \{\texttt{"lr"}, \texttt{"momentum"}\}. Prompts require \texttt{import torch} and \texttt{import torch.nn as nn} and instruct the model to output a single \texttt{nn.Module} definition without data loading or training loops. No pretrained weights or external feature extractors are permitted; only standard convolutional, pooling, normalization, and activation operations are allowed. An implicit edge-friendly latency target accompanies the $500{,}000$-parameter budget. The system message casts the model as an expert PyTorch architecture designer optimizing for first-epoch accuracy under these constraints; the user message specifies the dataset name, input/output shapes, parameter budget, and the API contract listed above. This prompt template is held fixed across all 22 cycles.

\section{Full Hyperparameter Specifications}
\label{app:hyperparams}

\paragraph{LoRA fine-tuning.}
Fine-tuning uses DeepSeek-Coder-7B-Instruct-v1.5 with LoRA applied to attention projections (q/k/v/o) and MLP projections (\texttt{up\_proj}, \texttt{down\_proj}, \texttt{gate\_proj}) in all 24 Transformer layers (0--23). LoRA hyperparameters are fixed across cycles: rank $r{=}32$, LoRA $\alpha{=}32$, dropout 0.05. The task is causal language modeling over chat-format prompt--response pairs. Each cycle runs for 5 epochs with learning rate $1\times10^{-5}$, per-device batch size 1, gradient accumulation 4 (effective batch size 4), paged AdamW in 8-bit, cosine schedule with 20 warmup steps, weight decay 0.01, max gradient norm 1.0, and \texttt{bfloat16} mixed precision.

\paragraph{Decoding configuration.}
Generation uses a chat interface with fixed decoding across all cycles: temperature 0.20, top-$k$ 50, nucleus $p{=}0.9$, \texttt{do\_sample=True}, and maximum new tokens 2{,}048 (padded with EOS). These settings are deliberately held constant so that changes in generator behavior are attributable solely to the evolving training corpus.

\section{Statistical Confidence Intervals}
\label{app:ci_formulas}

Let $N_{\mathrm{gen}}^{(c)}$ denote the number of candidate architectures sampled in cycle $c$, and $N_{\mathrm{valid}}^{(c)}$ the subset that compile and train successfully. The valid generation rate is $p_{\mathrm{valid}}^{(c)} = N_{\mathrm{valid}}^{(c)} / N_{\mathrm{gen}}^{(c)}$.

For all valid models in cycle $c$, let $\mathcal{A}^{(c)} = \{ A(m) : m \in \mathcal{M}_{\mathrm{valid}}^{(c)} \}$. For the sample mean $\bar{A}^{(c)}$ and standard deviation $s^{(c)}$ computed over $n_c = |\mathcal{A}^{(c)}|$, we report $t$-based 95\% confidence intervals:
\begin{equation}
    \bar{A}^{(c)} \pm t_{0.975,\,n_c - 1} \frac{s^{(c)}}{\sqrt{n_c}}.
\end{equation}
For any proportion $\hat{p} = k/n$ (e.g., $p_{\mathrm{valid}}^{(c)}$ or the proportion with $A(m) \geq \tau$), we report Wilson score confidence intervals~\citep{wilson1927probable}.

\section{Computational Complexity \& Accessibility}
To evaluate the practical accessibility of our framework, we benchmarked the computational requirements on consumer-grade hardware. Utilizing a single NVIDIA GeForce RTX 4090 (24GB), candidate generation averages 0.7--1.5 GPU-minutes per sample, while the low-fidelity proxy evaluation (1-epoch training) requires 1--5 GPU-minutes per valid model. Parameter-efficient fine-tuning via LoRA (rank 32, 5 epochs) over the terminal corpus of 2,153 pairs is completed in approximately 2--4 hours. We estimate the cumulative compute budget for the entire 22-cycle evolution at 90--266 GPU-hours, demonstrating that NNGPT enables sophisticated neural architecture design without requiring enterprise-scale clusters.

\section{Ethical Considerations (Extended)}
\label{app:ethics}

\paragraph{Data Usage and Privacy.}
The initial training corpus is derived from the publicly available LEMUR Neural Network Dataset. The dataset consists exclusively of source code and associated technical metadata and does not contain personally identifiable information (PII) or sensitive user data. Architectures generated during the iterative synthesis process are programmatically produced code artifacts within an isolated execution environment. As a result, no private, confidential, or user-generated data are introduced into the training or evaluation loop.

\paragraph{Security and Integrity of Generated Code.}
The LLM is used to generate executable source code as part of the architecture synthesis process. While syntactic and semantic validity checks are applied during experimentation, we note that any real-world deployment or reuse of LLM-generated code would necessitate comprehensive security reviews. Such audits would be required to mitigate risks related to unsafe coding practices or the inadvertent reproduction of vulnerable code patterns.

\paragraph{Transparency and Reproducibility.}
To support transparency and reproducibility, we provide a detailed account of the experimental setup, including the base LLM, fine-tuning strategy (LoRA), data filtering mechanisms (MinHash-Jaccard novelty filtering), and evaluation protocols. This level of documentation is intended to facilitate independent verification of the results and to enable replication or extension of the proposed methodology by the research community.

\end{document}